\documentclass[11pt,a4paper]{article}
\usepackage[hyperref]{acl2020}
\usepackage{times}
\usepackage{latexsym}
\usepackage{amsmath}
\usepackage{multirow}

\usepackage{enumitem} 
\usepackage{xfrac}

\usepackage{graphicx}

\usepackage[utf8]{inputenc}
\usepackage[T1]{fontenc}
\usepackage[english]{babel}

\usepackage{microtype}

\aclfinalcopy 

\usepackage{booktabs, tabularx}
\usepackage{amssymb}
\usepackage{subfig}

\usepackage{todonotes}

\DeclareMathSymbol{\shortminus}{\mathbin}{AMSa}{"39}

\title{Information-Theoretic Probing with Minimum Description Length}

  \author{Elena Voita$^{1,2}$\quad\quad\quad\quad\quad Ivan Titov$^{1,2}$\bigskip\\
  $^1$University of Edinburgh, Scotland  \\
  $^2$University of Amsterdam, Netherlands \\
  {\tt lena-voita@hotmail.com} \quad\quad {\tt ititov@inf.ed.ac.uk}
}

\date{}

\begin{document}
\maketitle
\begin{abstract}

To measure how well  pretrained representations 
encode some linguistic property, it is common to use accuracy of a probe, 
i.e. a classifier  trained to predict the  property from 
the representations. Despite widespread adoption of probes, differences in their accuracy fail to adequately reflect  differences in representations.  For example, they do not substantially favour pretrained representations
over randomly initialized ones. Analogously, their accuracy can be similar when probing for genuine linguistic labels and probing for random synthetic tasks. To see reasonable differences in accuracy with respect to these random baselines, previous work had to constrain either the amount of probe training data or its model size. 
Instead, we propose an alternative to the standard probes,
 information-theoretic probing with \textit{minimum description length} (MDL).  With MDL probing, training a probe to predict labels is recast as teaching it to effectively transmit the data. Therefore, the measure of interest changes from 
probe accuracy to the description length of labels given representations. In addition to probe quality, the description length evaluates `the amount of effort' needed to achieve the quality. This amount of effort characterizes either (i)~size of a probing model, or (ii)~the amount of data needed to achieve the high quality. We consider two methods for estimating MDL which can be easily implemented on top of the standard probing pipelines: variational coding and online coding. We show that these methods agree in results and are more informative and stable than the standard probes.\footnote{We release code at \url{https://github.com/lena-voita/description-length-probing}.}

\end{abstract}

\section{Introduction}

\begin{figure}[t!]
    \centering
     {\includegraphics[scale=0.38]{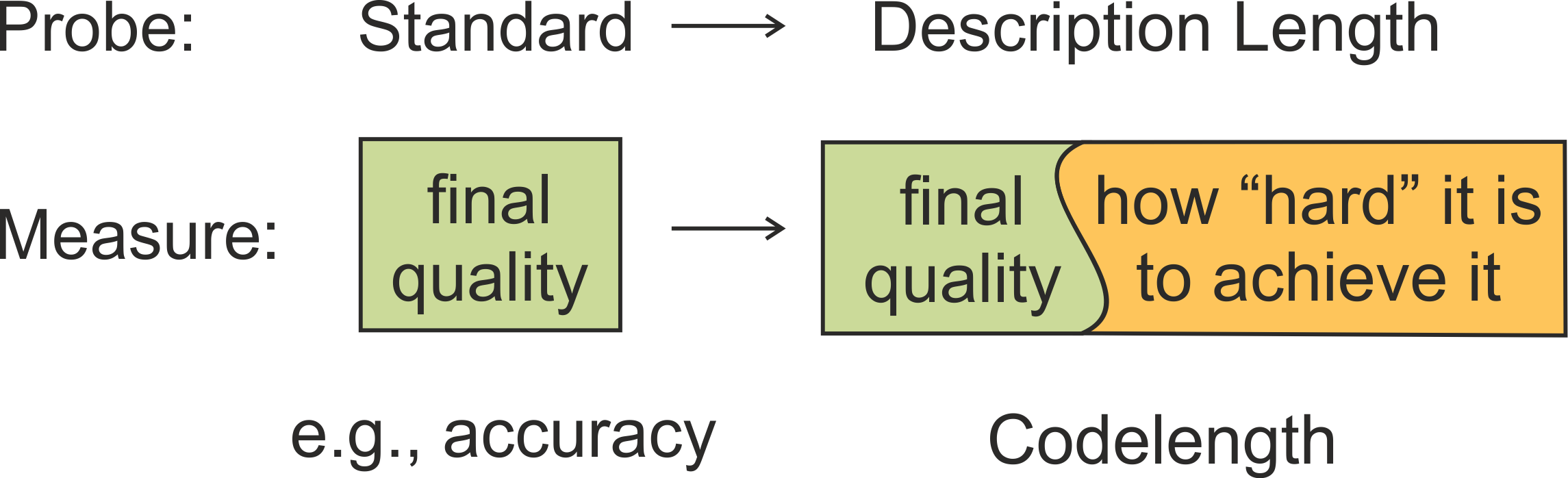}}
    \caption{Illustration of the idea behind MDL probes.}
    \label{fig:probe_main_intro}
\end{figure}

To estimate to what extent representations (e.g., ELMo~\cite{peters-etal-2018-deep} or BERT~\cite{devlin-etal-2019-bert}) capture a linguistic property, most previous work uses `probing tasks' (aka `probes' and `diagnostic classifiers'); 
see \citet{belinkov_survey} for a comprehensive review. These classifiers are trained to predict a linguistic property from `frozen' representations, and accuracy of the classifier is used to measure how well these representations encode the property.

Despite widespread adoption of such probes, 
they fail to adequately reflect differences in representations. This is clearly seen 
when using them to compare pretrained representations with randomly initialized ones~\cite{zhang-bowman-2018-language}. Analogously,
their accuracy can be similar when probing for genuine linguistic labels  and  probing  for  tags  randomly  associated to word types~(`control tasks',  \citet{hewitt-liang-2019-designing}). 
To see 
differences in the accuracy with respect to these {\it random baselines}, 
previous work had 
to reduce the amount of a probe training data~\cite{zhang-bowman-2018-language} or use smaller models for probes~\cite{hewitt-liang-2019-designing}.

As an alternative to the standard probing, we take an information-theoretic view at the task of measuring relations between representations and labels. Any regularity in representations with respect to labels can be exploited both to make predictions and to compress these labels, i.e., reduce length of the code needed to transmit them. Formally, we recast learning a model of data (i.e., training a probing classifier) as training it to transmit the data (i.e., labels) in as few bits as possible. This naturally leads to a change of measure: instead of evaluating  probe accuracy, we evaluate \textit{minimum description length~(MDL)} of labels given representations, i.e. the minimum number of bits needed to transmit the labels knowing the representations. Note that since labels are transmitted using a model, the model has to be transmitted as well (directly or indirectly). Thus, the overall codelength is a combination of the quality of fit of the model (compressed data length) with the cost of transmitting the model itself.

Intuitively, codelength characterizes not only the final quality of a probe, but also the `amount of effort' needed achieve this quality~(Figure~\ref{fig:probe_main_intro}). If representations have some clear structure with respect to labels, the relation between the representations and the labels can be understood with less effort; for example, (i)~the `rule' predicting the label 
(i.e., the probing model) can be simple, and/or (ii)~the amount of data needed to reveal this structure can be small. This is exactly how our vague (so far) notion of `the amount of effort' is translated into codelength. We explain this more formally when describing the two methods for evaluating MDL we use: \textit{variational coding} and \textit{online coding}; they differ in a way they incorporate model cost: directly or indirectly.

\textit{Variational code} explicitly incorporates cost of transmitting the model (probe weights) in addition to the cost of transmitting the labels; this joint cost is exactly the loss function of a variational learning algorithm~\cite{honkela_valpola}. As we will see in the experiments, close probe accuracies 
often come at a very different model cost: the `rule' (the probing model) explaining regularity in the data can be either simple (i.e., easy to communicate) or complicated (i.e., hard to communicate) depending on the strength of this regularity.

\textit{Online code} provides a way to transmit data without directly transmitting the model. Intuitively, it measures the ability to learn from different amounts of data. In this setting, the data is transmitted in a sequence of portions; at each step, the data transmitted so far is used to understand the regularity in this data and compress the following portion. If the regularity in the data is strong, it can be revealed using a small subset of the data, i.e., early in the transmission process, and can be exploited to efficiently transmit the rest of the dataset. The online code is related to the area under the learning curve, which plots quality as a function of the number of training examples.

If we now recall that, to get reasonable differences with random baselines,  previous work manually tuned (i) model size and/or (ii) the amount of data, we will see that these were 
indirect ways of accounting for the 
`amount of effort' component of (i) variational and (ii) online codes, respectively. Interestingly, since variational and online codes are different methods to estimate 
the same quantity  (and, as we will show, they agree in the results), we can conclude that
the ability of a probe to achieve good quality using a \textit{small amount of data} and its ability to achieve good quality using a \textit{small probe architecture} reflect the same property: 
\textit{strength of the regularity in the data}. In contrast to previous work, MDL incorporates this naturally in a theoretically justified way. Moreover, our experiments show that, differently from accuracy, conclusions made by MDL probes are not affected by an underlying probe setting, thus no manual search for settings is required.

We illustrate the effectiveness of MDL for different kinds of random baselines. For example, when considering  control tasks~\cite{hewitt-liang-2019-designing}, while probes have similar accuracies, these accuracies are achieved with a small probe model for the linguistic task and a large model for the random baseline (control task); these architectures are obtained as a byproduct of MDL optimization and not by manual search.

Our contributions are as follows:
\begin{itemize}
    \item we propose information-theoretic probing which measures MDL 
    of labels given representations;
    \item we show that MDL naturally characterizes not only probe quality, but also `the amount of effort' needed to achieve it; 
    \item we explain how to easily measure MDL on top of standard probe-training pipelines; 
    \item we show that results of MDL probing are more informative and stable than those of standard probes. 
\end{itemize}

\section{Information-Theoretic Viewpoint}

Let $\mathcal{D}=\{(x_1, y_1), (x_2, y_2), \dots, (x_n, y_n)\}$ be a dataset,
where  $x_{1:n}=(x_1, x_2, \dots, x_n)$ are representations from a model and $y_{1:n}=(y_1, y_2, \dots, y_n)$ are labels for some linguistic task (we assume that $y_i\in \{1, 2, \dots, K\}$, i.e. we consider classification tasks). As in standard probing task, we want to measure to what extent $x_{1:n}$ encode $y_{1:n}$. 
Differently from standard probes, we propose to look at this question from the information-theoretic perspective and define the goal of a probe as learning to effectively transmit the data.

\paragraph{Setting.} Following the standard information theory notation, let us imagine 
that Alice has all $(x_i, y_i)$ pairs in $\mathcal{D}$, Bob has just the $x_i$'s from $\mathcal{D}$, and that Alice wants to communicate the $y_i$'s to Bob. The task is to encode the labels $y_{1:n}$ knowing the inputs $x_{1:n}$ in an optimal way, i.e. with minimal codelength (in bits) needed to transmit $y_{1:n}$. 

\paragraph{Transmission: Data and Model.} Alice can transmit the labels using some probabilistic model of data $p(y|x)$ (e.g., it can be a trained probing classifier). Since Bob does not know the precise trained model that Alice is using, some explicit or implicit transmission of the model itself is also required.  In Section~\ref{sect:compression_bounds_def}, we explain how to transmit data using a model $p$. In Section~\ref{sect:compression_with_dl}, we show direct and indirect ways of transmitting the model.

\paragraph{Interpretation: quality and `amount of effort'.} In Section~\ref{sect:notes_after_code_description}, we show that total codelength characterizes both probe quality and the `amount of effort' needed to achieve it. We draw connections between different interpretations of this `amount of effort' part of the code and manual search for  probe settings done in previous work.\footnote{Note that in this work, we do not consider practical implementations of transmission algorithms; everywhere in the text,  `codelength' refers to the theoretical codelength of the associated encodings.}

\subsection{Transmission of Data Using a Model}
\label{sect:compression_bounds_def}

Suppose that Alice and Bob have agreed in advance on a model $p$, and both know the inputs $x_{1:n}$. Then there exists a code to transmit the labels $y_{1:n}$ losslessly with codelength\footnote{Up to at most one bit on the whole sequence; for datasets of reasonable size this can be ignored.} 
\vspace{-1ex}
\begin{equation}
L_{p}(y_{1:n}|x_{1:n})=-\sum\limits_{i=1}^n\log_2 p(y_i|x_i).
\label{eq:shannon_huffman}
\vspace{-0.8ex}
\end{equation}

This is the \textit{Shannon-Huffman code}, which gives an optimal bound on the codelength if the data are independent and come from a conditional probability distribution $p(y|x)$.

\paragraph{Learning is compression.} The bound~(\ref{eq:shannon_huffman}) is exactly the categorical cross-entropy loss evaluated on the model $p$. This shows that the task of compressing labels $y_{1:n}$ is equivalent to learning a model $p(y|x)$: 
quality of a learned model $p(y|x)$ is the codelength needed to transmit the data.

Compression is usually compared against \textit{uniform encoding} which does not require any learning from data. It assumes $p(y|x)=p_{unif}(y|x)=\frac{1}{K}$, and yields  codelength $L_{unif}(y_{1:n}|x_{1:n})=n\log_2K$ bits. 
 Another trivial encoding ignores input $x$ and relies on class priors $p(y)$, resulting in codelength
 $H(y)$.

\paragraph{Relation to Mutual Information.}
If the inputs and the outputs come from a true joint distribution $q(x, y)$, then, for any transmission method with codelength $L$, it holds $\mathbb{E}_q[L(y|x)] \ge H(y|x)
$~\cite{grunwald2004tutorial}. Therefore, the gain in codelength over the trivial codelength $H(y)$ is
$$
    H(y) - \mathbb{E}_q[L(y|x)] \le H(y) - H(y|x) = I(y; x).
$$
In other words, the compression is limited
by the mutual information (MI) between inputs (i.e. pretrained representations) and outputs (i.e. 
labels).

Note that total codelength includes model codelength in addition to the data code. This means that while high MI is necessary for effective compression, 
a good representation is the one which also yields 
simple models predicting $y$ from $x$, as we formalize in the next section.

\subsection{Transmission of the Model (Explicit or Implicit)}
\label{sect:compression_with_dl}

 We consider two compression methods that can be used with deep learning models (probing classifiers):
\begin{itemize}
    \item \textit{variational code}~-- an instance of \textit{two-part} codes, where a model is transmitted explicitly and then used to encode the data;
    \item \textit{online code}~-- a way to encode both model and data without \textit{directly} transmitting the model.
\end{itemize}

\subsubsection{Variational Code}

We assume that Alice and Bob have agreed on a model class $\mathcal{H} =\{p_{\theta} | \theta \in \Theta \}$.
With \textit{two-part} codes, 
for any model $p_{\theta^\ast}$, Alice first transmits its parameters $\theta^{\ast}$  and then encodes the data while relying on the model. The description length decomposes accordingly:
\begin{align}
\nonumber
& L^{2\shortminus part}_{\theta^{\ast}}(y_{1:n}|x_{1:n}) = \\ 
\nonumber
& = L_{param}(\theta^{\ast})
 + L_{p_{\theta^{\ast}}}(y_{1:n}|x_{1:n})  
 \\& = L_{param}(\theta^{\ast})
    -\sum\limits_{i=1}^{n}\log_2p_{\theta^{\ast}}(y_{i}|x_{i}). 
\end{align}
To compute the description length of the parameters $L_{param}(\theta^{\ast})$,
we can further assume that Alice and Bob have agreed on a prior distribution over the parameters $\alpha(\theta^{\ast})$. Now, we can rewrite the total description length as
\begin{align}
\nonumber
 & - \log_2 (\alpha(\theta^{\ast}) \epsilon^m)
    -\sum\limits_{i=1}^{n}\log_2p_{\theta^{\ast}}(y_{i}|x_{i}),
\end{align}
where  $m$ is the number of parameters and $\epsilon$
is a prearranged precision for each parameter. 
With deep learning models,
such straightforward codes for parameters 
are highly inefficient.
Instead, in the variational approach,
weights are treated as random variables,
and the description length is given by the expectation
\begin{align}
\nonumber
& L^{var}_{\beta}(y_{1:n}|x_{1:n}) =  \\
\nonumber
&
\!= \!   \shortminus \mathbb{E}_{\theta\sim\beta}\!\left[
    \log_2 \!\alpha(\theta) \! \shortminus\! \log_2 \!\beta (\theta)
   \! + \!\!\sum\limits_{i=1}^{n}\log_2p_{\theta}(y_{i}|x_{i})\!\right]\\
&\! =  KL(\beta\parallel\alpha) 
    \shortminus \mathbb{E}_{\theta\sim\beta} \sum\limits_{i=1}^{n}\log_2p_{\theta}(y_{i}|x_{i}),
    \label{eq:variational_code} 
\end{align}
where $\beta(\theta) = \prod_{i=1}^{m}{ \beta_i(\theta_i)}$ is a distribution encoding uncertainty about the parameter values. The distribution $\beta(\theta)$ is chosen by minimizing the codelength given in Expression~(\ref{eq:variational_code}).
The formal justification for the description length relies on the {\it bits-back} argument~\cite{hinton93keeping,honkela_valpola,mackay2003information}. However, the underlying intuition is straightforward:
parameters we are uncertain about can be transmitted
at a lower cost as
the uncertainty
 can be used to determine the required 
precision. The entropy term in Equation~(\ref{eq:variational_code}),
$H(\beta) =\shortminus\mathbb{E}_{\theta\sim\beta} \log_2 \beta (\theta)$,
quantifies this discount.

 The negated codelength $- L^{var}_{\beta}(y_{1:n}|x_{1:n})$
is known as the evidence-lower-bound (ELBO) and used as the objective in variational inference. The distribution $\beta(\theta)$ approximates the intractable posterior distribution $p(\theta | x_{1:n}, y_{1:n})$. Consequently, any variational method can in principle be used to estimate the  codelength.  

In our experiments, we use the network compression method of~\citet{louizos_bayesian}. Similarly to variational dropout~\cite{molchanov2017variational},
it uses sparsity-inducing priors on the parameters, pruning neurons from the probing classifier as a byproduct of optimizing the ELBO. As a result we can assess the  probe complexity both using its description length $KL(\beta \parallel \alpha)$ and by inspecting the  discovered architecture.

\subsubsection{Online (or Prequential) Code}

The online (or prequential) code~\cite{rissanen1984universal} is a way to encode both the model and the labels without directly encoding the model weights. In the online setting, Alice and Bob agree on the form of the model $p_{\theta}(y|x)$ with learnable parameters $\theta$, its initial random seeds, and its learning algorithm. They also choose timesteps $1 = t_0 < t_1 < \dots < t_S = n$ and encode data by blocks.\footnote{In all experiments in this paper, the timesteps  correspond to 0.1, 0.2, 0.4, 0.8, 1.6, 3.2, 6.25, 12.5, 25, 50, 100 percent of the dataset.} Alice starts by communicating $y_{1:t_1}$ with a uniform code, then both Alice and Bob learn a model $p_{\theta_{1}}(y|x)$ that predicts $y$ from $x$ using data $\{(x_i, y_i)\}_{i=1}^{t_1}$, and Alice uses that model to communicate the next data block $y_{t_1+1:t_2}$. Then both Alice and Bob learn a model $p_{\theta_{2}}(y|x)$ from a larger block $\{(x_i, y_i)\}_{i=1}^{t_2}$ and use it to encode $y_{t_2+1:t_3}$. This process continues until the entire dataset has been transmitted. The resulting online codelength is
\begin{align}
\nonumber
& L^{online}(y_{1:n}|x_{1:n}) = t_1\log_2K -  \\
& -\sum\limits_{i=1}^{S-1}\log_2p_{\theta_i}(y_{t_i+1:t_{i+1}}|x_{t_i+1:t_{i+1}}).
\label{eq:online_code}
\end{align}

In this sequential evaluation, a model that performs well with a limited number of training examples will be rewarded by having a shorter codelength (Alice will require fewer bits to transmit the subsequent $y_{t_i:t_{i+1}}$ to Bob). The online code is related to the area under the learning curve, which plots quality (in case of probes, accuracy) as a function of the number of training examples. We will illustrate this in Section~\ref{sect:control_exp_results_pos}.

\subsection{Interpretations of Codelength}
\label{sect:notes_after_code_description}

\paragraph{Connection to previous work.} To get larger differences in scores compared to random baselines, previous work tried to (i)~reduce size of a probing model and (ii)~reduce the amount of a probe training data. Now we can see that these were indirect ways to account for the `amount of effort' component of (i) variational and (ii) online codes, respectively.

\paragraph{Online code and model size.} While the online code does not incorporate model cost explicitly, we can still evaluate model cost by interpreting the difference between the cross-entropy of the model trained on all data and online codelength as the cost of the model.
The former is codelength of the data if one knows model parameters, the latter~(online codelength) --- if one does not know them. In Section~\ref{sect:control_exp_results_pos} we will show that trends for model cost evaluated for the online code are similar to those for the variational code. It means that in terms of a code, the ability of a probe to achieve good quality using \textit{small amount of data} or 
using \textit{a small probe architecture} reflect the same property:
 the \textit{strength of the regularity in the data}.

\paragraph{Which code to choose?}  In terms of implementation, the online code uses a standard probe along with its training setting: it trains the  probe on increasing subsets of the dataset. Using the variational code requires changing (i)~a probing model to a Bayesian model and (ii)~the loss function to the corresponding variational loss~(\ref{eq:variational_code}) (i.e. adding the model $KL$ term to the standard data cross-entropy).
As we will show later, these methods agree in results. Therefore, the choice of the method can be done depending on the preferences: the variational code can be used to inspect the induced probe architecture, but the online code is easier to implement.

\begin{table*}[h!t!]
\centering
\begin{tabular}{lccc}
\toprule
 & Labels & Number of sentences & Number of targets \\
\midrule
Part-of-speech & 45 & 39832 / 1700 / 2416 & 950028 / 40117 / 56684 \\
\bottomrule
\end{tabular}
\vspace{-1ex}
\caption{Dataset statistics. Numbers of sentences and targets are given for train / dev / test sets.}
\label{tab:control_dataset_stat}
\end{table*}

\begin{table*}[t!]
\centering
\begin{tabular}{lcccccc}
\toprule
 & & \bf Accuracy & \multicolumn{4}{c}{\bf Description Length}\\
 & &   & \multicolumn{2}{c}{\bf variational code} & \multicolumn{2}{c}{\bf online code}  \\
 & &    &  codelength & compression & codelength & compression \\
\midrule
\multicolumn{5}{l}{\!\!\!\bf MLP-2, h=1000}\\
& \textsc{layer} 0 & 93.7 / 96.3 & 163 / 267 & 31{.}32 / 19{.}09  & 173 / 302 & 29.5 / 16.87 \\
& \textsc{layer 1} & 97.5 / 91.9 & 85 / 470 & 59{.}76 / 10{.}85 & 96 / 515 & 53.06 / 9.89 \\
& \textsc{layer 2} & 97.3 / 89.4 & 103 / 612 & 49{.}67 / 8{.}33  & 115 / 717 & 44.3 / 7.11  \\

\bottomrule
\end{tabular}
\caption{Experimental results; shown in pairs: linguistic task / control task. Codelength is measured in kbits (variational codelength is given in equation (\ref{eq:variational_code}), online -- in equation (\ref{eq:online_code})). Accuracy is shown for the standard probe as in~\citet{hewitt-liang-2019-designing}; for the variational probe, scores are similar (see Table~\ref{tab:control_bayes_pruned_architecture}).}
\label{tab:control_main_results}
\vspace{-1ex}
\end{table*}

\section{Description Length and Control Tasks}

\citet{hewitt-liang-2019-designing} noted that probe accuracy itself does not necessarily reveal if the representations encode the linguistic annotation or if the probe `itself' learned to predict this annotation.
They introduced control tasks which associate word types with random outputs, and each word token is assigned its type's output, regardless of context. By construction, such tasks can only be learned by the probe itself.  They argue that selectivity, i.e. difference between linguistic task accuracy and control task accuracy, reveals how much the linguistic probe relies on the 
regularities encoded in the representations. They propose to tune probe hyperparameters so that to maximize selectivity. 
In contrast, we will show that MDL probes do not require such tuning.

\subsection{Experimental Setting}

In all experiments, we use the data and follow the setting of \citet{hewitt-liang-2019-designing}; we build on top of their code
and release our extended version to reproduce the experiments.

In the main text, we use a probe with default hyperparameters which was a starting point in~\citet{hewitt-liang-2019-designing} and was shown to have low selectivity. In the appendix, we provide results for 10 different settings and show that, in contrast to accuracy, codelength is stable across settings. 

\paragraph{Task: part of speech.} Control tasks were designed for two tasks: part-of-speech (PoS) tagging and dependency edge prediction. In this work, we focus only on the PoS tagging task, 
the 
task of assigning tags, such as noun, verb, and adjective, to individual word tokens. 
For the control task, for each word type, a PoS tag is independently sampled from the empirical distribution of the tags in the linguistic data.

\paragraph{Data.} The pretrained model is the 5.5 billion-word pre-trained ELMo~\cite{peters-etal-2018-deep}. The data comes from Penn Treebank~\cite{marcus-etal-1993-building} with the traditional  parsing training/development/testing splits\footnote{As given by the code of~\citet{qi-manning-2017-arc} at \url{https://github.com/qipeng/arc-swift}.} without extra preprocessing. 
Table~\ref{tab:control_dataset_stat} shows dataset statistics.

\paragraph{Probes.}
The probe is MLP-2 of \citet{hewitt-liang-2019-designing} with the default hyperparameters. Namely,
it is a multi-layer perceptron with two hidden layers defined as: $y_i \sim $ softmax$(W_3 ReLU(W_2 ReLU(W_1h_i)))$;  hidden layer size $h$ is 1000 and no dropout is used. Additionally, in the appendix, we provide results for both MLP-2 and MLP-1 for several $h$ values: 1000, 500, 250, 100, 50.

For the variational code, we replace dense layers with the Bayesian compression layers from~\citet{louizos_bayesian}; the loss function changes to Eq.~(\ref{eq:variational_code}). 

\paragraph{Optimizer.} All of our probing models are trained with Adam~\cite{adam-optimizer} with learning rate 0.001. With standard probes, we follow the original paper~\cite{hewitt-liang-2019-designing} and anneal the learning  rate by a factor of 0.5 once the epoch does not lead to a new minimum loss on the development set; we stop training when 4 such epochs occur in a row. With variational probes, we do not anneal learning rate and train probes for 200 epochs; long training is recommended to enable  pruning~\cite{louizos_bayesian}.

\begin{figure*}[t!]
    \centering
    \subfloat[]
    {\includegraphics[scale=0.28]{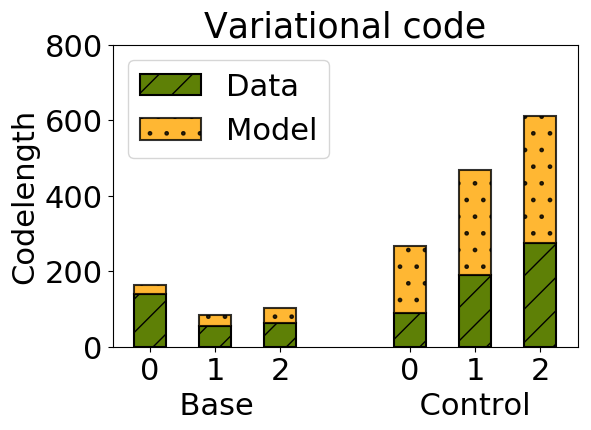}}
    \subfloat[]
    {\includegraphics[scale=0.28]{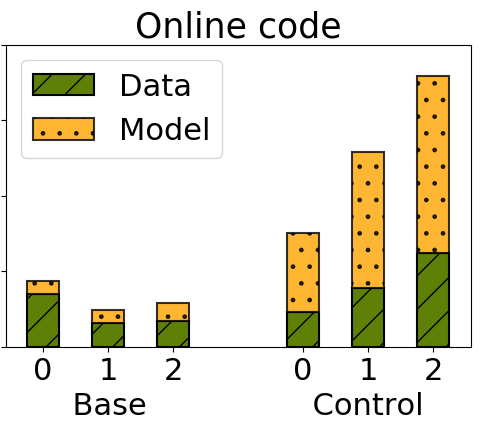}}
    \quad
    \subfloat[]
    {\includegraphics[scale=0.18]{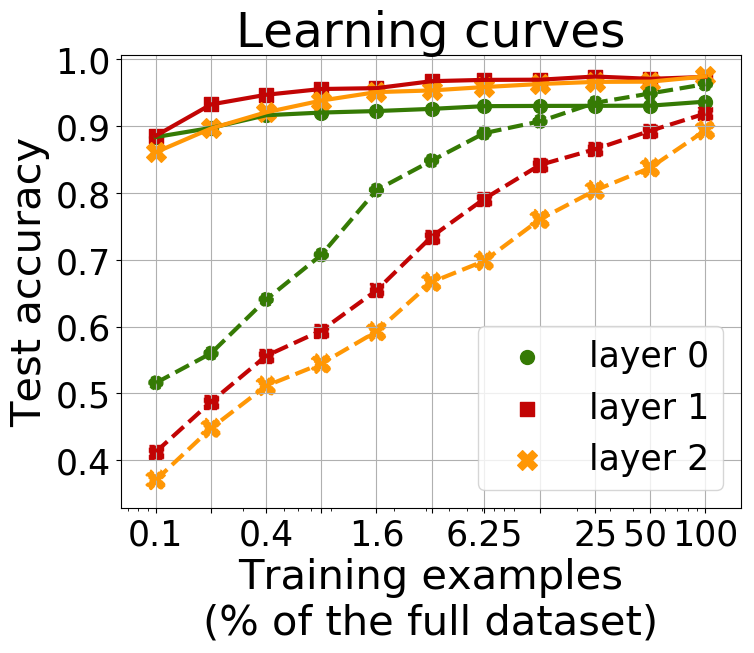}}  
    \quad
    \subfloat[random seeds]
    {\includegraphics[scale=0.36]{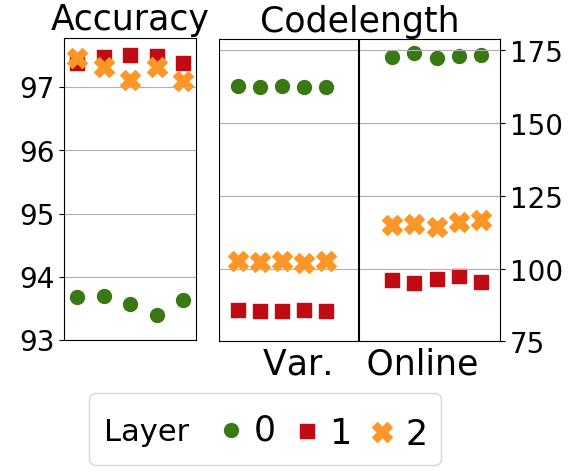}}
    \vspace{-1ex}
    \caption{(a), (b): codelength split into data and model codes; (c): learning curves corresponding to online code (solid lines for linguistic task, dashed -- for control); (d): results for 5 random seeds, linguistic task (for control task, see appendix).}
    \label{fig:by_type_control_pos}
    \vspace{-1ex}
\end{figure*}

\subsection{Experimental Results}
\label{sect:control_exp_results_pos}

Results are shown in Table~\ref{tab:control_main_results}.\footnote{Accuracies can differ from the ones reported in~\citet{hewitt-liang-2019-designing}: we report accuracy on the test set, while they -- on the development
set. Since the development set is used for stopping criteria, we believe that test scores are more reliable.}

\paragraph{Different compression methods, similar results.}
First, we see that both compression methods show similar trends in codelength.
For the linguistic task, the best layer is the first one. For the control task, codes become larger as we move up from the embedding layer; this is expected since the control task measures the ability to memorize word type. Note that codelengths for control tasks are substantially larger than for the linguistic task (at least twice larger). This again illustrates that description length is preferable to probe accuracy: in contrast to accuracy, codelength is able to distinguish these tasks without any search for settings.

\paragraph{\textsc{Layer 0}: MDL is correct, accuracy is not.}
What is even more surprising, \textit{codelength identifies the control task even when accuracy indicates the opposite}: for \textsc{layer 0}, accuracy for the control task is higher, but 
the code is twice longer than for the linguistic task.
This is because codelength characterizes how hard it is to achieve this accuracy: for the control task, accuracy is higher, but the cost of achieving this score is very big. We will illustrate this later in this section.

\paragraph{Embedding vs contextual: drastic difference.} For the linguistic task, note that codelength for the embedding layer is approximately twice larger than that for the first layer. 
Later in Section~\ref{sect:mdl_and_random_models} we will see the same trends for several other tasks, and will show that even contextualized representations obtained with a randomly initialized model are a lot better than with the embedding layer alone.

\paragraph{Model: small for linguistic, large for control.} Figure~\ref{fig:by_type_control_pos}(a) shows data and model components of the variational code. For control tasks, model size is several times larger than for the linguistic task. This is something that probe accuracy alone is not able to reflect: representations have structure with respect to the linguistic labels and this structure can be `explained' with a small model. The same representations do not have structure with respect to random labels, therefore these labels can be predicted only using a larger model.

Using interpretation from Section~\ref{sect:notes_after_code_description} to split the online code into data and model codelength, we get Figure~\ref{fig:by_type_control_pos}(b). The trends are similar to the ones with the variational code; but with the online code, the model component shows how easy it is to learn from small amount of data: if the representations have structure with respect to some labels, this structure can be revealed with a few training examples. Figure~\ref{fig:by_type_control_pos}(c) shows learning curves showing the difference between behavior of the linguistic and control tasks. In addition to probe accuracy, such learning curves have also been used by \citet{yogatama2019learning} and \citet{talmor2019olmpics}.

\begin{table}[t!]
\centering
\begin{tabular}{lccc}
\toprule
& & \bf Accuracy & \bf Final probe \\
\midrule

\multicolumn{2}{l}{\!\!\!\bf layer 0}\\
& \!\!\!\!\!\!base & 93.5 & 406-33-49 \\
& \!\!\!\!\!\!control & 96.3 & 427-214-137 \\
\midrule

\multicolumn{2}{l}{\!\!\!\bf layer 1}\\
& \!\!\!\!\!\!base & 97.7 & 664-55-35 \\
& \!\!\!\!\!\!control & 92.2 & 824-272-260 \\
\midrule

\multicolumn{2}{l}{\!\!\!\bf layer 2}\\
& \!\!\!\!\!\!base & 97.3 & 750-75-41 \\
& \!\!\!\!\!\!control & 88.7 & 815-308-481 \\
\bottomrule
\end{tabular}
\caption{Pruned architecture of a trained variational probe (starting probe: 1024-1000-1000).}
\label{tab:control_bayes_pruned_architecture}
\end{table}

\begin{table*}[t!]
\centering
\begin{tabular}{ll}
\toprule
Part-of-speech &  \textcolor{gray}{I want to find more ,} [something]  \textcolor{gray}{bigger or deeper .} $\longrightarrow$ NN (Noun) \\
Constituents &\textcolor{gray}{I want to find more ,} [something bigger or deeper] \textcolor{gray}{.} $\longrightarrow$ NP (Noun Phrase) \\
Dependencies & [I]$_1$ \textcolor{gray}{am not} [sure]$_2$ \textcolor{gray}{how reliable that is , though .} $\longrightarrow$ nsubj (nominal subject)  \\
Entities & \textcolor{gray}{The most fascinating is the maze known as} [Wind Cave] \textcolor{gray}{.} $\longrightarrow$ LOC \\
SRL & \textcolor{gray}{I want to} [find]$_{1}$ [more , something bigger or deeper]$_{2}$ \textcolor{gray}{.} $\longrightarrow$ Agr1 (Agent) \\
Coreference &  \textcolor{gray}{So} [the followers]$_1$ \textcolor{gray}{waited to say anything about what} [they]$_2$ \textcolor{gray}{saw .} $\longrightarrow$ True\\
Rel. (SemEval) & \textcolor{gray}{The} [shaman]$_1$ \textcolor{gray}{cured him with} [herbs]$_2$ \textcolor{gray}{ .} $\longrightarrow$ Instrument-Agency(e2, e1)\\
\bottomrule
\end{tabular}
\caption{Examples of sentences, spans, and target labels for each task.}
\label{tab:jiant_datasets_examples}
\end{table*}


\begin{table*}[t!]
\centering
\begin{tabular}{lccc}
\toprule
 & Labels & Number of sentences & Number of targets \\
\midrule
Part-of-speech & 48 & 115812 / 15680 / 12217 & 2070382 / 290013 / 212121 \\
Constituents & 30 & 115812 / 15680 / 12217 & 1851590 / 255133 / 190535 \\
Dependencies & 49 & 12522 / 2000 / 2075 & 203919 / 25110 / 25049 \\
Entities & 18 & 115812 / 15680 / 12217 & 128738 / 20354 / 12586 \\
SRL & 66 & 253070 / 35297 / 26715 & 598983 / 83362 / 61716 \\
Coreference & 2 & 115812 / 15680 / 12217 & 207830 / 26333 / 27800 \\
Rel. (SemEval) & 19 & 6851 / 1149 / 2717 & 6851 / 1149 / 2717 \\
\bottomrule
\end{tabular}
\caption{Dataset statistics. Numbers of sentences and targets are given for train / dev / test sets.}
\label{tab:jiant_datasets_stat}
\end{table*}

\paragraph{Architecture: sparse for linguistic, dense for control.} 
The method for the variational code we use, Bayesian compression of~\citet{louizos_bayesian}, lets us assess the induced probe complexity not only by using its description length (as we did above), but also by looking at the induced architecture (Table~\ref{tab:control_bayes_pruned_architecture}). Probes learned for linguistic tasks are much smaller than those for control tasks, with only 33-75 neurons at the second and third layers. 
This relates to previous work~\cite{hewitt-liang-2019-designing}. The authors considered several predefined probe architectures and picked one of them based on a manually defined criterion. In contrast, the variational code gives probe architecture as a byproduct of training and does not need human guidance.

\begin{figure}[t!]
    \centering
     {\includegraphics[scale=0.28]{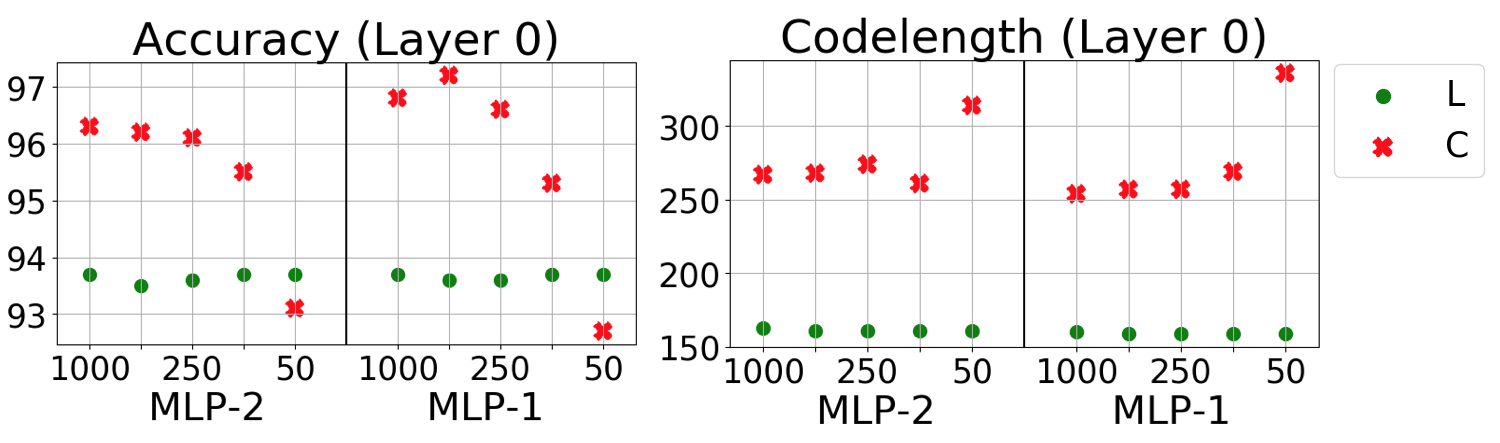}}
    \caption{Results for 10 probe settings: accuracy is wrong for 8 out of 10 settings, MDL is always correct (for accuracy higher is better, for codelength -- lower). }
    \label{fig:control_settings}
\end{figure}

\subsection{Stability and Reliability of MDL Probes}
\label{sect:control_stability}

Here we discuss stability of MDL results across compression methods, underlying probing classifier setting and random seeds.

\paragraph{The two compression methods agree in results.}
Note that the observed agreement in codelengths from different methods (Table~\ref{tab:control_main_results}) is rather surprising: this contrasts to \citet{blier_ollivier_desc_length}, who experimented with images (MNIST, CIFAR-10) and argued that the variational code yields very poor compression bounds compared to online code. We can speculate that their results may be due to the particular 
variational approach they use. 
The agreement between different codes is desirable and suggests sensibility and reliability of the results.

\paragraph{Hyperparameters: change results for accuracy, do not for MDL.} While here we will discuss in detail results for the default settings, in the appendix we provide results for 10 different settings; for \textsc{layer 0}, results are given in Figure~\ref{fig:control_settings}. 
We see that accuracy can change greatly with the settings.  For example,  difference in accuracy for linguistic and control tasks varies a lot; for \textsc{layer 0} there are settings with contradictory results: accuracy can be higher either for the linguistic or for the control task depending on the settings (Figure~\ref{fig:control_settings}). In striking contrast to accuracy, \textit{MDL results are stable across settings}, thus \textit{MDL does not require search for probe settings}.

\paragraph{Random seed: affects accuracy but not MDL.} We evaluated results from Table~\ref{tab:control_main_results} for random seeds from 0 to 4; for the linguistic task, results are shown in Figure~\ref{fig:by_type_control_pos}(d). We see that using accuracy can lead to different rankings of layers depending on a random seed, making it hard to draw conclusions about their relative qualities. For example, accuracy for \textsc{layer 1} and \textsc{layer 2} are 97.48 and 97.31 for seed 1, but 97.38 and 97.48 for seed 0.  On the contrary, the MDL results are stable and the scores given to different layers are well separated.

Note that for this `real' task, where the true ranking of layers 1 and 2 is not known in advance, tuning a probe setting by maximizing difference with the synthetic control task (as done by~\citet{hewitt-liang-2019-designing}) does not help: in the tuned setting, scores for these layers remain very close (e.g., 97.3 and 97.0~\cite{hewitt-liang-2019-designing}).

\section{Description Length and Random Models}
\label{sect:mdl_and_random_models}

Now, from random labels for word types, we come to another type of random baselines: randomly initialized models. Probes using these representations show surprisingly strong performance for both token~\cite{zhang-bowman-2018-language} and sentence~\cite{wieting2018no} representations. This again confirms that accuracy alone does not reflect what a representation encodes.  With MDL probes, we will see that codelength shows large difference between trained and randomly initialized representations.

In this part, we also experiment with ELMo and compare it with a version of the ELMo model in which all weights above the lexical layer (\textsc{layer 0}) are replaced with random orthonormal matrices (but the embedding layer, \textsc{layer 0}, is retained from trained ELMo). 
We conduct a line of experiments using a suite of edge probing tasks~\cite{tenney2018you}. In these tasks, a probing model (Figure~\ref{fig:edge_probing}) can access only representations within given spans, such as a predicate-argument pair, and must predict properties, such as semantic roles.

\subsection{Experimental Setting}

\begin{table*}[t!]
\begin{minipage}{0.7\textwidth}
\centering
\setlength\tabcolsep{1.5 pt}
\begin{tabular}{lcccccc}
\toprule
 & \textbf{\ \ \ \ \ \ }& \bf{\ \ \ \ Accuracy \ \ } & \multicolumn{4}{c}{\bf Description Length}\\
 & &   & \multicolumn{2}{c}{\bf variational code}  & \multicolumn{2}{c}{\bf online code} \\
 & &    & codelength \ \ & compression & codelength & compression \\
\midrule
\multicolumn{5}{l}{\!\bf Part-of-speech}\\
& L0 & 91.3 & 483 & 23.4  & 462 & 24.5  \\
& L1 & 97.8 / 95.7 &  209 / 273 & 54.0 / 41.4  & 192 / 294 & 58.8 / 38.5 \!\!\!\!\\
& L2 & 97.5 / 95.7 & 252 / 273 & 44.7 / 41.4  & 216 / 294 & 52.3 / 38.5  \!\!\!\!\\

\midrule

\multicolumn{5}{l}{\!\bf Constituents}\\
& L0 & 75.9 &  1181 & 7.5  & 1149  & 7.7   \\
& L1 & 86.4 / 77.6 &  603 / 877 & 14.7 / 10.1  & 570 / 1081 & 15.6 / 8.2  \!\!\!\!\\
& L2 & 85.1 / 77.6 &  719 / 875 & 12.3 / 10.1  & 680 / 1074 & 13.1 / 8.3  \!\!\!\!\\

\midrule

\multicolumn{5}{l}{\!\bf Dependencies}\\
& L0 & 80.9 &  158  & 7.1  & 175  & 6.4   \\
& L1 & 94.0 / 90.3 &  80 / 103 & 14.0 / 10.8  & 74 / 106 & 15.1 / 10.5  \!\!\!\!\\
& L2 & 92.8 / 90.4 &  94 / 103 & 11.9 / 10.8  & 82 / 106 & 13.7 / 10.6  \!\!\!\!\\

\midrule

\multicolumn{5}{l}{\!\bf Entities}\\
& L0 &  92.3  &  40  & 13.2    &  40 & 13.1  \\
& L1 & 95.0 / 93.5 &   27 / 34 & 19.3 / 15.4   & 27 / 35 & 19.8 / 15.1  \!\!\!\!\\
& L2 & 95.3 / 93.6 &   30 / 34 & 17.7 / 15.2   & 26 / 35 & 19.9 / 15.1  \!\!\!\!\\
\midrule

\multicolumn{5}{l}{\!\bf SRL}\\
& L0 & 81.1   &  411  & 8.6   & 381  & 9.3   \\
& L1 &   91.9 / 84.4   & 228 / 306 & 15.5 / 11.5   & 212 / 365 & 16.7 / 9.7  \!\!\!\!\\
& L2 &  90.2 / 84.5  & 272 / 306 & 13.0 / 11.6   & 245 / 363 & 14.4 / 9.7  \!\!\!\!\\

\midrule

\multicolumn{5}{l}{\!\bf Coreference}\\
& L0 & 89.9  & 57.4  & 3.54   & 60 & 3.4   \\
& L1 & 92.9 / 90.7 &  50.3 / 54.5 & 4.04 / 3.72  & 51 / 65 & 4.0 / 3.1  \\
& L2 & 92.2 / 90.4 &  56.8 / 54.3 & 3.57 / 3.74  & 55 / 65 & 3.7 / 3.1  \\

\midrule

\multicolumn{5}{l}{\!\bf Rel. (SemEval)}\\
& L0 & 55.8  &  11.5  & 2.48    &  15.9  & 1.79    \\
& L1 & 75.2 / 69.1 &  8.0 / 9.7 & 3.56 / 2.94   & 8.8 / 11.8 & 3.2 / 2.4  \\
& L2 & 77.0 / 68.9  &  8.4 / 9.7 & 3.40 / 2.92   &  8.6 / 11.7 & 3.3 / 2.4  \\

\bottomrule
\end{tabular}
\caption{Experimental results; shown in pairs: trained model / randomly initialized model. Codelength is measured in kbits  (variational codelength is given in equation (\ref{eq:variational_code}), online -- in equation (\ref{eq:online_code})), compression -- with respect to the corresponding uniform code.}
\label{tab:jiant_main}
\end{minipage}\hspace{10px}
\begin{minipage}{0.215\textwidth}
    \centering
    {\includegraphics[scale=0.25]{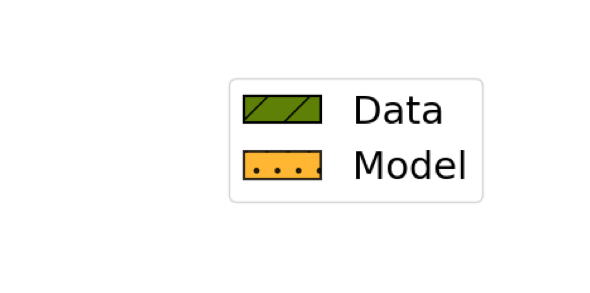}}
    
    {\includegraphics[scale=0.215]{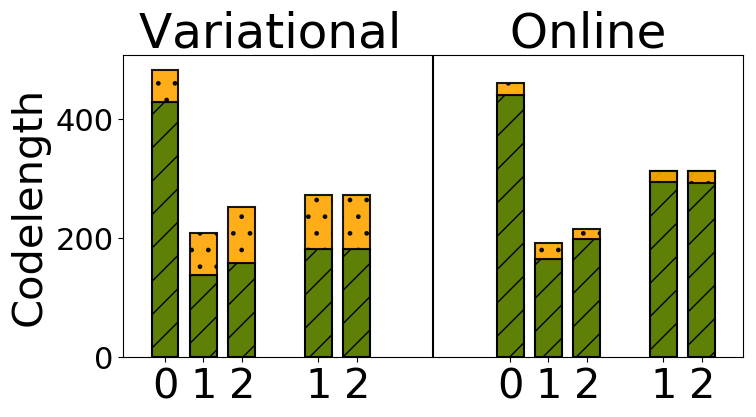}}
    
    {\includegraphics[scale=0.215]{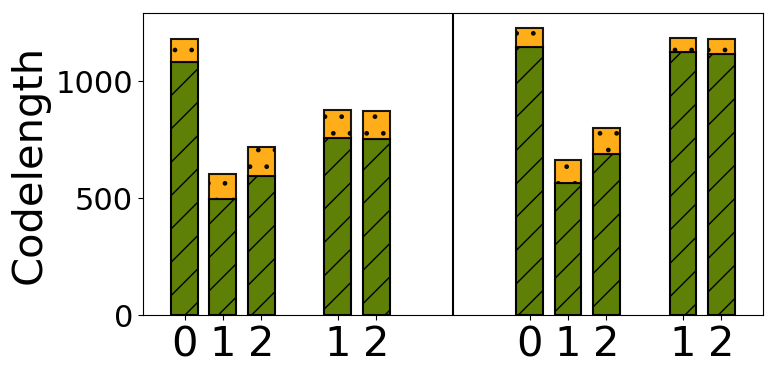}}
    
    {\includegraphics[scale=0.215]{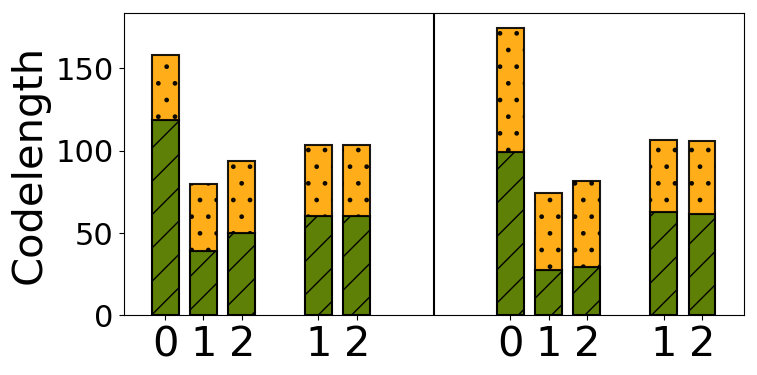}}
    
    {\includegraphics[scale=0.215]{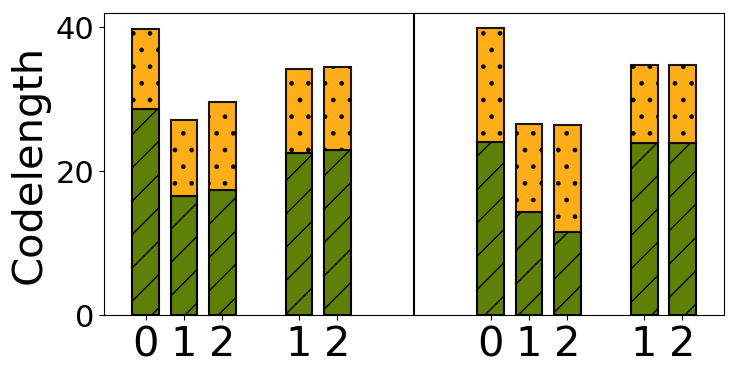}}
    
    {\includegraphics[scale=0.215]{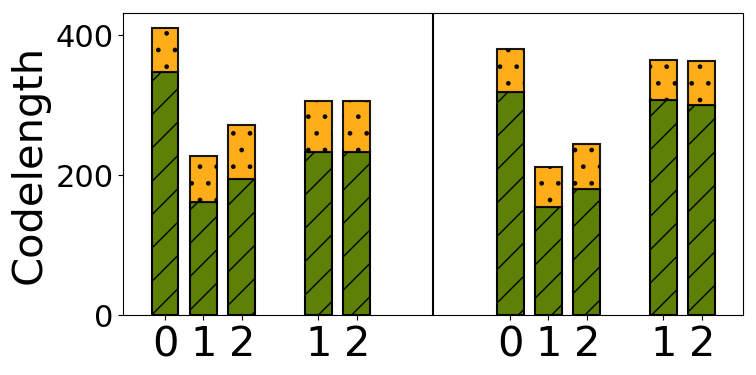}}
    
    {\includegraphics[scale=0.215]{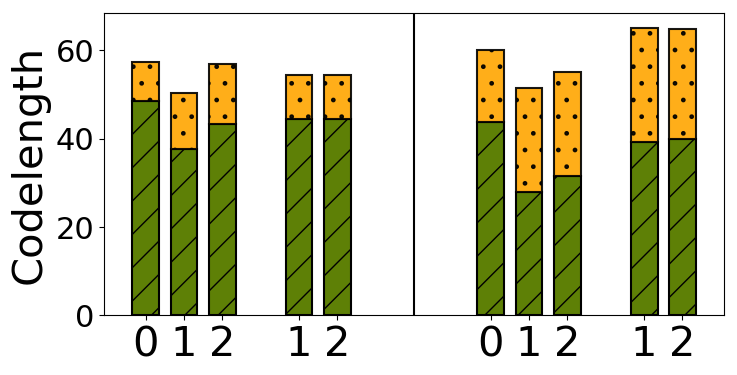}}
    
    {\includegraphics[scale=0.215]{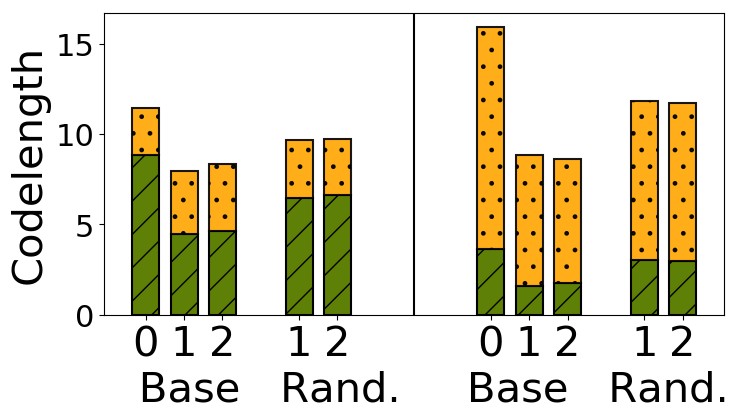}}

    \caption{Data and model code components for the tasks from Table~\ref{tab:jiant_main}.}
    \label{fig:by_type_jiant}

\end{minipage}
\end{table*}

\paragraph{Tasks and datasets.} We focus on several core NLP tasks: PoS tagging, syntactic constituent and dependency labeling,  
named entity recognition, semantic role labeling, coreference resolution, and relation classification. Examples for each task are shown in Table~\ref{tab:jiant_datasets_examples}, dataset statistics are in Table~\ref{tab:jiant_datasets_stat}. See extra details in~\citet{tenney2018you}.

\begin{figure}[t!]
    \centering
     {\includegraphics[scale=0.6]{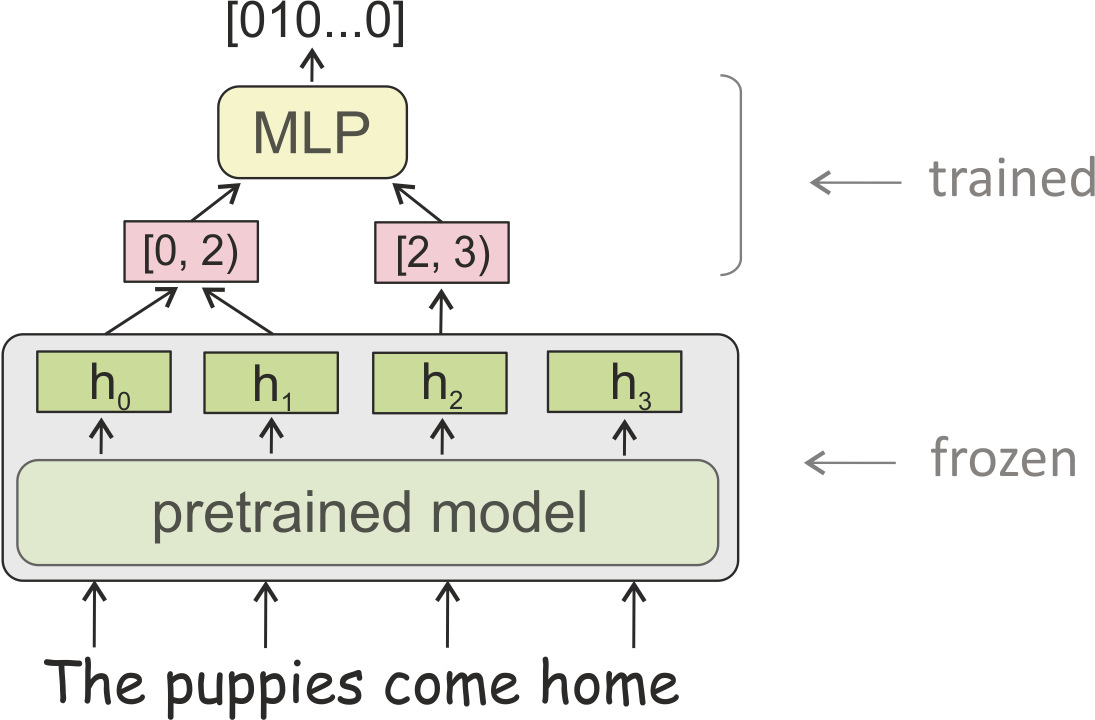}}
    \caption{Probing model architecture for an edge probing task. The example is for semantic role labeling; for PoS, NER and constituents, only a single span is used.}
    \label{fig:edge_probing}
\end{figure}

We follow \citet{tenney2018you} and use ELMo~\cite{peters-etal-2018-deep} trained on the Billion Word Benchmark dataset~\cite{chelba2014one}.

\paragraph{Probes and optimization.} Probing architecture is illustrated in Figure~\ref{fig:edge_probing}. It takes a list of contextual vectors $[e_0, e_1, \dots, e_n]$ and integer spans $s_1 = [i_1, j_1)$ and (optionally) $s_2 = [i_2, j_2)$ as inputs, and uses a projection layer followed by the self-attention pooling operator of~\citet{lee-etal-2017-end} to compute fixed-length span representations. The span representations are concatenated and fed into a two-layer MLP followed by a softmax output layer. As in the original paper, we use the standard cross-entropy loss, hidden layer size of 256 and dropout of 0{.}3. For further details on training, we refer the reader to the original paper by~\citet{tenney2018you}.\footnote{The differences with the original implementation by~\citet{tenney2018you} are: softmax with the cross-entropy loss instead of sigmoid with binary cross-entropy, using the loss instead of F1 in the early stopping criterion.}

For the variational code, the layers are replaced with that of Bayesian compression by~\citet{louizos_bayesian}; loss function changes to~(\ref{eq:variational_code})
and no dropout is used. Similar to the experiments in the previous section, we do not anneal learning rate and train at least 200 epochs to enable  pruning. 

We build our experiments on top of the original code by~\citet{tenney2018you} and release our extended version.

\subsection{Experimental Results}

Results are shown in Table~\ref{tab:jiant_main}. 

\paragraph{\textsc{Layer 0} vs contextual.}
As we have already seen in the previous section, codelength shows drastic difference between the embedding layer (\textsc{layer 0}) and contextualized representations: codelengths differ about twice for most of the tasks. Both compression methods show that even for the randomly initialized model, contextualized representations are better than lexical representations. 
This is because context-agnostic embeddings do not contain enough information about the task, i.e., MI between labels and context-agnostic representations is smaller than between labels and contextualized representations. Since compression of the labels given model (i.e., data component of the code) is limited by the MI between the representations and the labels (Section~\ref{sect:compression_bounds_def}), the data component of the codelength is much bigger for the embedding layer than for contextualized representations. 

\paragraph{Trained vs random.}
As expected, codelengths for the randomly initialized model are larger than for the trained one. 
This is more prominent when not just looking at the bare scores, but comparing compression against context-agnostic representations. For all tasks, compression bounds for the randomly initialized model are closer to those of context-agnostic \textsc{Layer 0} than representations from the trained model. This shows that gain from using context for the randomly initialized model is at least twice smaller than for the trained model.

Note also that \textit{randomly initialized layers do not evolve}: for all tasks, MDL  for layers of the randomly initialized model is the same. Moreover, Table~\ref{fig:by_type_jiant} shows that not only total codelength but data and model components of the code for random model layers are also identical. 
For the trained model, this is not the case: \textsc{layer 2} is worse than \textsc{layer 1} for all tasks.  This is one more illustration of the general process explained in~\citet{voita-etal-2019-bottom}: the way representations evolve between layers is defined by the training objective. For the randomly initialized model, since no training objective has been optimized, no evolution happens.

\section{Related work }

Probing classifiers are the most common approach for associating neural network representations with linguistic properties~(see \citet{belinkov_survey} for a survey). Among the works highlighting limitations of standard probes (not mentioned earlier) is the work by
\citet{saphra-lopez-2019-understanding}, who show that diagnostic classifiers are not suitable for understanding learning dynamics. 

In addition to task performance, learning curves have also been used before by \citet{yogatama2019learning} to evaluate how quickly a model learns a new task,  and by \citet{talmor2019olmpics} to understand whether the performance of a LM on a task should be attributed to the pre-trained representations or to the process of fine-tuning on the task data.

Other methods for analyzing NLP models include (i) inspecting the mechanisms a model uses to encode information, such as attention weights~\cite{voita-etal-2018-context,raganato-tiedemann:2018:BlackboxNLP,voita-etal-2019-analyzing,clark-etal-2019-bert,kovaleva-etal-2019-revealing} or individual neurons~\cite{karpathy2015visualizing,pham-etal-2016-convolutional,bau2019neurons-in-mt}, (ii) looking at model predictions using manually defined templates, either evaluating sensitivity to specific grammatical errors~\cite{linzen-Q16-1037,colorless-green,tran-etal-2018-importance,marvin-linzen-2018-targeted} or understanding what language models know when applying them as knowledge bases or in question answering settings~\cite{radford2019language,petroni-etal-2019-language,poerner2019bert,jiang2019can}. 

An  information-theoretic view on analysis of NLP models has been previously attempted in~\citet{voita-etal-2019-bottom} when explaining how representations in the Transformer evolve between layers under different training objectives.

\section{Conclusions}

We propose information-theoretic probing which measures \textit{minimum description length~(MDL)} of labels given representations. We show that MDL naturally characterizes not only probe quality, but also `the amount of effort' needed to achieve it (or, intuitively, \textit{strength of the regularity} in representations with respect to the labels); this is done in a  theoretically justified way without manual search for settings. We explain how to easily measure MDL on top of standard probe-training pipelines. We show that results of
MDL probing are more informative and stable compared to the standard probes.

\section*{Acknowledgments}
IT acknowledges support of the European Research Council (ERC StG BroadSem 678254) and the Dutch National Science Foundation (NWO VIDI 639.022.518). 

\bibliography{acl2020}

\begin{thebibliography}{40}
\expandafter\ifx\csname natexlab\endcsname\relax\def\natexlab#1{#1}\fi

\bibitem[{Bau et~al.(2019)Bau, Belinkov, Sajjad, Durrani, Dalvi, and
  Glass}]{bau2019neurons-in-mt}
Anthony Bau, Yonatan Belinkov, Hassan Sajjad, Nadir Durrani, Fahim Dalvi, and
  James Glass. 2019.
\newblock \href {https://openreview.net/pdf?id=H1z-PsR5KX} {Identifying and
  controlling important neurons in neural machine translation}.
\newblock In \emph{International Conference on Learning Representations}, New
  Orleans.

\bibitem[{Belinkov and Glass(2019)}]{belinkov_survey}
Yonatan Belinkov and James Glass. 2019.
\newblock \href {https://doi.org/10.1162/tacl\_a\_00254} {Analysis methods in
  neural language processing: A survey}.
\newblock \emph{Transactions of the Association for Computational Linguistics},
  7:49--72.

\bibitem[{Blier and Ollivier(2018)}]{blier_ollivier_desc_length}
L\'{e}onard Blier and Yann Ollivier. 2018.
\newblock \href
  {http://papers.nips.cc/paper/7490-the-description-length-of-deep-learning-models.pdf}
  {The description length of deep learning models}.
\newblock In \emph{Advances in Neural Information Processing Systems}, pages
  2216--2226.

\bibitem[{Chelba et~al.(2014)Chelba, Mikolov, Schuster, Ge, Brants, Koehn, and
  Robinson}]{chelba2014one}
Ciprian Chelba, Tomas Mikolov, Mike Schuster, Qi~Ge, Thorsten Brants, Phillipp
  Koehn, and Tony Robinson. 2014.
\newblock One billion word benchmark for measuring progress in statistical
  language modeling.
\newblock In \emph{Fifteenth Annual Conference of the International Speech
  Communication Association}.

\bibitem[{Clark et~al.(2019)Clark, Khandelwal, Levy, and
  Manning}]{clark-etal-2019-bert}
Kevin Clark, Urvashi Khandelwal, Omer Levy, and Christopher~D. Manning. 2019.
\newblock \href {https://doi.org/10.18653/v1/W19-4828} {What does {BERT} look
  at? an analysis of {BERT}{'}s attention}.
\newblock In \emph{Proceedings of the 2019 ACL Workshop BlackboxNLP: Analyzing
  and Interpreting Neural Networks for NLP}, pages 276--286, Florence, Italy.
  Association for Computational Linguistics.

\bibitem[{Devlin et~al.(2019)Devlin, Chang, Lee, and
  Toutanova}]{devlin-etal-2019-bert}
Jacob Devlin, Ming-Wei Chang, Kenton Lee, and Kristina Toutanova. 2019.
\newblock \href {https://doi.org/10.18653/v1/N19-1423} {{BERT}: Pre-training of
  deep bidirectional transformers for language understanding}.
\newblock In \emph{Proceedings of the 2019 Conference of the North {A}merican
  Chapter of the Association for Computational Linguistics: Human Language
  Technologies, Volume 1 (Long and Short Papers)}, pages 4171--4186,
  Minneapolis, Minnesota. Association for Computational Linguistics.

\bibitem[{Grunwald(2004)}]{grunwald2004tutorial}
Peter Grunwald. 2004.
\newblock A tutorial introduction to the minimum description length principle.
\newblock \emph{arXiv preprint math/0406077}.

\bibitem[{Gulordava et~al.(2018)Gulordava, Bojanowski, Grave, Linzen, and
  Baroni}]{colorless-green}
Kristina Gulordava, Piotr Bojanowski, Edouard Grave, Tal Linzen, and Marco
  Baroni. 2018.
\newblock \href {https://doi.org/10.18653/v1/N18-1108} {Colorless green
  recurrent networks dream hierarchically}.
\newblock In \emph{Proceedings of the 2018 Conference of the North American
  Chapter of the Association for Computational Linguistics: Human Language
  Technologies, Volume 1 (Long Papers)}, pages 1195--1205. Association for
  Computational Linguistics.

\bibitem[{Hewitt and Liang(2019)}]{hewitt-liang-2019-designing}
John Hewitt and Percy Liang. 2019.
\newblock \href {https://doi.org/10.18653/v1/D19-1275} {Designing and
  interpreting probes with control tasks}.
\newblock In \emph{Proceedings of the 2019 Conference on Empirical Methods in
  Natural Language Processing and the 9th International Joint Conference on
  Natural Language Processing (EMNLP-IJCNLP)}, pages 2733--2743, Hong Kong,
  China. Association for Computational Linguistics.

\bibitem[{Hinton and von Cramp(1993)}]{hinton93keeping}
GE~Hinton and D~von Cramp. 1993.
\newblock Keeping neural networks simple by minimising the description length
  of weights.
\newblock In \emph{Proceedings of COLT-93}, pages 5--13.

\bibitem[{{Honkela} and {Valpola}(2004)}]{honkela_valpola}
Antti {Honkela} and Harri {Valpola}. 2004.
\newblock \href {https://doi.org/10.1109/TNN.2004.828762} {Variational learning
  and bits-back coding: an information-theoretic view to bayesian learning}.
\newblock In \emph{IEEE Transactions on Neural Networks}, volume~15, pages
  800--810.

\bibitem[{Jiang et~al.(2019)Jiang, Xu, Araki, and Neubig}]{jiang2019can}
Zhengbao Jiang, Frank~F Xu, Jun Araki, and Graham Neubig. 2019.
\newblock How can we know what language models know?
\newblock \emph{arXiv preprint arXiv:1911.12543}.

\bibitem[{Karpathy et~al.(2015)Karpathy, Johnson, and
  Fei-Fei}]{karpathy2015visualizing}
Andrej Karpathy, Justin Johnson, and Li~Fei-Fei. 2015.
\newblock Visualizing and understanding recurrent networks.
\newblock \emph{arXiv preprint arXiv:1506.02078}.

\bibitem[{Kingma and Ba(2015)}]{adam-optimizer}
Diederik Kingma and Jimmy Ba. 2015.
\newblock {Adam: A method for stochastic optimization}.
\newblock In \emph{{Proceedings of the {International} {Conference} on
  {Learning} {Representation} (ICLR 2015)}}.

\bibitem[{Kovaleva et~al.(2019)Kovaleva, Romanov, Rogers, and
  Rumshisky}]{kovaleva-etal-2019-revealing}
Olga Kovaleva, Alexey Romanov, Anna Rogers, and Anna Rumshisky. 2019.
\newblock \href {https://doi.org/10.18653/v1/D19-1445} {Revealing the dark
  secrets of {BERT}}.
\newblock In \emph{Proceedings of the 2019 Conference on Empirical Methods in
  Natural Language Processing and the 9th International Joint Conference on
  Natural Language Processing (EMNLP-IJCNLP)}, pages 4365--4374, Hong Kong,
  China. Association for Computational Linguistics.

\bibitem[{Lee et~al.(2017)Lee, He, Lewis, and Zettlemoyer}]{lee-etal-2017-end}
Kenton Lee, Luheng He, Mike Lewis, and Luke Zettlemoyer. 2017.
\newblock \href {https://doi.org/10.18653/v1/D17-1018} {End-to-end neural
  coreference resolution}.
\newblock In \emph{Proceedings of the 2017 Conference on Empirical Methods in
  Natural Language Processing}, pages 188--197, Copenhagen, Denmark.
  Association for Computational Linguistics.

\bibitem[{Linzen et~al.(2016)Linzen, Dupoux, and Goldberg}]{linzen-Q16-1037}
Tal Linzen, Emmanuel Dupoux, and Yoav Goldberg. 2016.
\newblock \href {http://aclweb.org/anthology/Q16-1037} {Assessing the ability
  of lstms to learn syntax-sensitive dependencies}.
\newblock \emph{Transactions of the Association for Computational Linguistics},
  4:521--535.

\bibitem[{Louizos et~al.(2017)Louizos, Ullrich, and Welling}]{louizos_bayesian}
Christos Louizos, Karen Ullrich, and Max Welling. 2017.
\newblock \href
  {http://papers.nips.cc/paper/6921-bayesian-compression-for-deep-learning.pdf}
  {Bayesian compression for deep learning}.
\newblock In \emph{Advances in Neural Information Processing Systems}, pages
  3288--3298.

\bibitem[{MacKay(2003)}]{mackay2003information}
David~JC MacKay. 2003.
\newblock \emph{Information theory, inference and learning algorithms}.
\newblock Cambridge university press.

\bibitem[{Marcus et~al.(1993)Marcus, Santorini, and
  Marcinkiewicz}]{marcus-etal-1993-building}
Mitchell~P. Marcus, Beatrice Santorini, and Mary~Ann Marcinkiewicz. 1993.
\newblock \href {https://www.aclweb.org/anthology/J93-2004} {Building a large
  annotated corpus of {E}nglish: The {P}enn {T}reebank}.
\newblock \emph{Computational Linguistics}, 19(2):313--330.

\bibitem[{Marvin and Linzen(2018)}]{marvin-linzen-2018-targeted}
Rebecca Marvin and Tal Linzen. 2018.
\newblock \href {https://doi.org/10.18653/v1/D18-1151} {Targeted syntactic
  evaluation of language models}.
\newblock In \emph{Proceedings of the 2018 Conference on Empirical Methods in
  Natural Language Processing}, pages 1192--1202, Brussels, Belgium.
  Association for Computational Linguistics.

\bibitem[{Molchanov et~al.(2017)Molchanov, Ashukha, and
  Vetrov}]{molchanov2017variational}
Dmitry Molchanov, Arsenii Ashukha, and Dmitry Vetrov. 2017.
\newblock Variational dropout sparsifies deep neural networks.
\newblock In \emph{Proceedings of the 34th International Conference on Machine
  Learning}.

\bibitem[{Peters et~al.(2018)Peters, Neumann, Iyyer, Gardner, Clark, Lee, and
  Zettlemoyer}]{peters-etal-2018-deep}
Matthew Peters, Mark Neumann, Mohit Iyyer, Matt Gardner, Christopher Clark,
  Kenton Lee, and Luke Zettlemoyer. 2018.
\newblock \href {https://doi.org/10.18653/v1/N18-1202} {Deep contextualized
  word representations}.
\newblock In \emph{Proceedings of the 2018 Conference of the North {A}merican
  Chapter of the Association for Computational Linguistics: Human Language
  Technologies, Volume 1 (Long Papers)}, pages 2227--2237, New Orleans,
  Louisiana. Association for Computational Linguistics.

\bibitem[{Petroni et~al.(2019)Petroni, Rockt{\"a}schel, Riedel, Lewis, Bakhtin,
  Wu, and Miller}]{petroni-etal-2019-language}
Fabio Petroni, Tim Rockt{\"a}schel, Sebastian Riedel, Patrick Lewis, Anton
  Bakhtin, Yuxiang Wu, and Alexander Miller. 2019.
\newblock \href {https://doi.org/10.18653/v1/D19-1250} {Language models as
  knowledge bases?}
\newblock In \emph{Proceedings of the 2019 Conference on Empirical Methods in
  Natural Language Processing and the 9th International Joint Conference on
  Natural Language Processing (EMNLP-IJCNLP)}, pages 2463--2473, Hong Kong,
  China. Association for Computational Linguistics.

\bibitem[{Pham et~al.(2016)Pham, Kruszewski, and
  Boleda}]{pham-etal-2016-convolutional}
Ngoc-Quan Pham, German Kruszewski, and Gemma Boleda. 2016.
\newblock \href {https://doi.org/10.18653/v1/D16-1123} {Convolutional neural
  network language models}.
\newblock In \emph{Proceedings of the 2016 Conference on Empirical Methods in
  Natural Language Processing}, pages 1153--1162, Austin, Texas. Association
  for Computational Linguistics.

\bibitem[{Poerner et~al.(2019)Poerner, Waltinger, and
  Sch{\"u}tze}]{poerner2019bert}
Nina Poerner, Ulli Waltinger, and Hinrich Sch{\"u}tze. 2019.
\newblock Bert is not a knowledge base (yet): Factual knowledge vs. name-based
  reasoning in unsupervised qa.
\newblock \emph{arXiv preprint arXiv:1911.03681}.

\bibitem[{Qi and Manning(2017)}]{qi-manning-2017-arc}
Peng Qi and Christopher~D. Manning. 2017.
\newblock \href {https://doi.org/10.18653/v1/P17-2018} {Arc-swift: A novel
  transition system for dependency parsing}.
\newblock In \emph{Proceedings of the 55th Annual Meeting of the Association
  for Computational Linguistics (Volume 2: Short Papers)}, pages 110--117,
  Vancouver, Canada. Association for Computational Linguistics.

\bibitem[{Radford et~al.(2019)Radford, Wu, Child, Luan, Amodei, and
  Sutskever}]{radford2019language}
Alec Radford, Jeffrey Wu, Rewon Child, David Luan, Dario Amodei, and Ilya
  Sutskever. 2019.
\newblock Language models are unsupervised multitask learners.
\newblock \emph{OpenAI Blog}, 1(8):9.

\bibitem[{Raganato and Tiedemann(2018)}]{raganato-tiedemann:2018:BlackboxNLP}
Alessandro Raganato and Jörg Tiedemann. 2018.
\newblock \href {http://www.aclweb.org/anthology/W18-5431} {An analysis of
  encoder representations in transformer-based machine translation}.
\newblock In \emph{Proceedings of the 2018 EMNLP Workshop BlackboxNLP:
  Analyzing and Interpreting Neural Networks for NLP}, pages 287--297,
  Brussels, Belgium. Association for Computational Linguistics.

\bibitem[{Rissanen(1984)}]{rissanen1984universal}
Jorma Rissanen. 1984.
\newblock Universal coding, information, prediction, and estimation.
\newblock \emph{IEEE Transactions on Information theory}, 30(4):629--636.

\bibitem[{Saphra and Lopez(2019)}]{saphra-lopez-2019-understanding}
Naomi Saphra and Adam Lopez. 2019.
\newblock \href {https://doi.org/10.18653/v1/N19-1329} {Understanding learning
  dynamics of language models with {SVCCA}}.
\newblock In \emph{Proceedings of the 2019 Conference of the North {A}merican
  Chapter of the Association for Computational Linguistics: Human Language
  Technologies, Volume 1 (Long and Short Papers)}, pages 3257--3267,
  Minneapolis, Minnesota. Association for Computational Linguistics.

\bibitem[{Talmor et~al.(2019)Talmor, Elazar, Goldberg, and
  Berant}]{talmor2019olmpics}
Alon Talmor, Yanai Elazar, Yoav Goldberg, and Jonathan Berant. 2019.
\newblock \href {https://arxiv.org/abs/1912.13283} {olmpics -- on what language
  model pre-training captures}.
\newblock \emph{arXiv preprint arXiv:1912.13283}.

\bibitem[{Tenney et~al.(2019)Tenney, Xia, Chen, Wang, Poliak, McCoy, Kim,
  Van~Durme, Bowman, Das et~al.}]{tenney2018you}
Ian Tenney, Patrick Xia, Berlin Chen, Alex Wang, Adam Poliak, R~Thomas McCoy,
  Najoung Kim, Benjamin Van~Durme, Samuel~R Bowman, Dipanjan Das, et~al. 2019.
\newblock \href {https://openreview.net/pdf?id=SJzSgnRcKX} {What do you learn
  from context? probing for sentence structure in contextualized word
  representations}.
\newblock In \emph{International Conference on Learning Representations}.

\bibitem[{Tran et~al.(2018)Tran, Bisazza, and Monz}]{tran-etal-2018-importance}
Ke~Tran, Arianna Bisazza, and Christof Monz. 2018.
\newblock \href {https://doi.org/10.18653/v1/D18-1503} {The importance of being
  recurrent for modeling hierarchical structure}.
\newblock In \emph{Proceedings of the 2018 Conference on Empirical Methods in
  Natural Language Processing}, pages 4731--4736, Brussels, Belgium.
  Association for Computational Linguistics.

\bibitem[{Voita et~al.(2019{\natexlab{a}})Voita, Sennrich, and
  Titov}]{voita-etal-2019-bottom}
Elena Voita, Rico Sennrich, and Ivan Titov. 2019{\natexlab{a}}.
\newblock \href {https://doi.org/10.18653/v1/D19-1448} {The bottom-up evolution
  of representations in the transformer: A study with machine translation and
  language modeling objectives}.
\newblock In \emph{Proceedings of the 2019 Conference on Empirical Methods in
  Natural Language Processing and the 9th International Joint Conference on
  Natural Language Processing (EMNLP-IJCNLP)}, pages 4396--4406, Hong Kong,
  China. Association for Computational Linguistics.

\bibitem[{Voita et~al.(2018)Voita, Serdyukov, Sennrich, and
  Titov}]{voita-etal-2018-context}
Elena Voita, Pavel Serdyukov, Rico Sennrich, and Ivan Titov. 2018.
\newblock \href {https://doi.org/10.18653/v1/P18-1117} {Context-aware neural
  machine translation learns anaphora resolution}.
\newblock In \emph{Proceedings of the 56th Annual Meeting of the Association
  for Computational Linguistics (Volume 1: Long Papers)}, pages 1264--1274,
  Melbourne, Australia. Association for Computational Linguistics.

\bibitem[{Voita et~al.(2019{\natexlab{b}})Voita, Talbot, Moiseev, Sennrich, and
  Titov}]{voita-etal-2019-analyzing}
Elena Voita, David Talbot, Fedor Moiseev, Rico Sennrich, and Ivan Titov.
  2019{\natexlab{b}}.
\newblock \href {https://doi.org/10.18653/v1/P19-1580} {Analyzing multi-head
  self-attention: Specialized heads do the heavy lifting, the rest can be
  pruned}.
\newblock In \emph{Proceedings of the 57th Annual Meeting of the Association
  for Computational Linguistics}, pages 5797--5808, Florence, Italy.
  Association for Computational Linguistics.

\bibitem[{Wieting and Kiela(2019)}]{wieting2018no}
John Wieting and Douwe Kiela. 2019.
\newblock \href {https://openreview.net/forum?id=BkgPajAcY7} {No training
  required: Exploring random encoders for sentence classification}.
\newblock In \emph{International Conference on Learning Representations}.

\bibitem[{Yogatama et~al.(2019)Yogatama, de~Masson~d'Autume, Connor, Kocisky,
  Chrzanowski, Kong, Lazaridou, Ling, Yu, Dyer, and
  Blunsom}]{yogatama2019learning}
Dani Yogatama, Cyprien de~Masson~d'Autume, Jerome Connor, Tomas Kocisky, Mike
  Chrzanowski, Lingpeng Kong, Angeliki Lazaridou, Wang Ling, Lei Yu, Chris
  Dyer, and Phil Blunsom. 2019.
\newblock \href {https://arxiv.org/abs/1901.11373} {Learning and evaluating
  general linguistic intelligence}.
\newblock \emph{arXiv preprint arXiv:1901.11373}.

\bibitem[{Zhang and Bowman(2018)}]{zhang-bowman-2018-language}
Kelly Zhang and Samuel Bowman. 2018.
\newblock \href {https://www.aclweb.org/anthology/W18-5448} {Language modeling
  teaches you more than translation does: Lessons learned through auxiliary
  syntactic task analysis}.
\newblock In \emph{Proceedings of the 2018 {EMNLP} Workshop {B}lackbox{NLP}:
  Analyzing and Interpreting Neural Networks for {NLP}}, pages 359--361,
  Brussels, Belgium. Association for Computational Linguistics.

\end{thebibliography}
\bibliographystyle{acl_natbib}

\appendix

\section{Description Length and Control Tasks}

\subsection{Settings}

Results are given in Table~\ref{tab:control_results_settings}.

\newcommand{\spacelena}{\!\!\!\!\!\!}

\begin{table}[h!]
\small
\centering
\begin{tabular}{lccccc}
\toprule
  & \!\!\!\!\!\!\!\!\bf Accuracy \!\!\!\!\!\!\!\!\ & \multicolumn{4}{c}{\spacelena\bf Description Length}\\
  &   & \multicolumn{2}{c}{\bf variational code} & \multicolumn{2}{c}{\spacelena\bf online code}  \\
  &    & \spacelena codelength &\spacelena compr. & \spacelena codelength &\spacelena compr. \spacelena\\
\midrule
\multicolumn{5}{l}{\!\!\!\bf MLP-2, h=1000}\\
\!\!\!\textsc{l} 0 &\spacelena 93.7 / 96.3 &\spacelena 163 / 267 &\spacelena 32 / 19  &\spacelena 173 / 302 &\spacelena 30 / 17 \spacelena\\
\!\!\!\textsc{l 1} &\spacelena 97.5 / 91.9 &\spacelena 85 / 470 &\spacelena 60 / 11 &\spacelena 96 / 515 &\spacelena 53 / 10 \spacelena\\
\!\!\!\textsc{l 2} &\spacelena 97.3 / 89.4 &\spacelena 103 / 612 &\spacelena 50 / 8  &\spacelena 115 / 717 &\spacelena 44 / 7 \spacelena \\

\midrule
\multicolumn{5}{l}{\!\!\!\bf MLP-2, h=500}\\
\!\!\!\textsc{l 0} &\spacelena 93.5 / 96.2 & \spacelena161 / 268 &\spacelena 32 / 19 &\spacelena 170 / 313 &\spacelena 30 / 16 \\
\!\!\!\textsc{l 1} &\spacelena 97.8 / 92.1 & \spacelena84 / 470 &\spacelena 61 / 11 &\spacelena 93 / 547 &\spacelena 55 / 9 \\
\!\!\!\textsc{l 2} &\spacelena 97.1 / 86.5 &\spacelena 102 / 611 &\spacelena 50 / 8 &\spacelena 112 / 755 &\spacelena 46 / 7 \\

\midrule
\multicolumn{5}{l}{\!\!\!\bf MLP-2, h=250}\\
\!\!\!\textsc{l 0} &\spacelena 93.6 / 96.1 &\spacelena 161 / 274 &\spacelena 32 / 19 &\spacelena 169 / 328 & \spacelena30 / 16 \\
\!\!\!\textsc{l 1} &\spacelena 97.7 / 90.3 &\spacelena 84 / 470 &\spacelena 61 / 11 &\spacelena 91 / 582 &\spacelena 56 / 9 \\
\!\!\!\textsc{l 2} &\spacelena 97.1 / 85.2 &\spacelena 101 / 611 &\spacelena 50 / 8 &\spacelena 112 / 799 &\spacelena 46 / 6 \\

\midrule
\multicolumn{5}{l}{\!\!\!\bf MLP-2, h=100 }\\
\!\!\!\textsc{l 0} &\spacelena 93.7 / 95.5 & \spacelena161 / 261 &\spacelena 32 / 20 &\spacelena 167 / 367 &\spacelena 31 / 14 \\
\!\!\!\textsc{l 1} &\spacelena 97.6 / 86.9 &\spacelena 84 / 492 &\spacelena 61 / 10 & \spacelena91 / 678 &\spacelena 56 / 8 \\
\!\!\!\textsc{l 2} &\spacelena 97.2 / 80.9 & \spacelena102 / 679 &\spacelena 50 / 8 &\spacelena 112 / 901 &\spacelena 46 / 6 \\

\midrule
\multicolumn{5}{l}{\!\!\!\bf MLP-2, h=50}\\
\!\!\!\textsc{l 0} &\spacelena 93.7 / 93.1 &\spacelena 161 / 314 &\spacelena 32 / 16 &\spacelena 166 / 416 &\spacelena 31 / 12 \\
\!\!\!\textsc{l 1} &\spacelena 97.6 / 82.7 & \spacelena84 / 605 &\spacelena 61 / 8 &\spacelena 93 / 781 &\spacelena 55 / 7 \\
\!\!\!\textsc{l 2} &\spacelena 97.0 / 76.2 &\spacelena 102 / 833 &\spacelena 50 / 6 &\spacelena 116 / 1007 &\spacelena 44 / 5 \\

\midrule
\multicolumn{5}{l}{\!\!\!\bf MLP-1, h=1000}\\
\!\!\!\textsc{l 0} &\spacelena 93.7 / 96.8 & \spacelena160 / 254 &\spacelena 32 / 20 &\spacelena 166 / 275 &\spacelena 31 / 19 \\
\!\!\!\textsc{l 1} &\spacelena 97.7 / 92.7 & \spacelena82 / 468 &\spacelena 62 / 11 &\spacelena 88 / 477 &\spacelena 58 / 11 \\
\!\!\!\textsc{l 2} &\spacelena 97.0 / 86.7 & \spacelena100 / 618 &\spacelena 51 / 8 &\spacelena 107 / 696 &\spacelena 48 / 7 \\

\midrule
\multicolumn{5}{l}{\!\!\!\bf MLP-1, h=500}\\
\!\!\!\textsc{l 0} &\spacelena 93.6 / 97.2 &\spacelena 159 / 257 &\spacelena 32 / 20 &\spacelena 164 / 295 &\spacelena 31 / 17 \\
\!\!\!\textsc{l 1} &\spacelena 97.5 / 91.6 & \spacelena82 / 468 &\spacelena 62 / 11 &\spacelena 88 / 516 &\spacelena 58 / 10 \\
\!\!\!\textsc{l 2} &\spacelena 97.0 / 86.3 &\spacelena 100 / 619 &\spacelena 51 / 8 &\spacelena 107 / 736 &\spacelena 48 / 7 \\

\midrule
\multicolumn{5}{l}{\!\!\!\bf MLP-1, h=250}\\
\!\!\!\textsc{l 0} & \spacelena 93.6 / 96.6 &\spacelena 159 / 257 &\spacelena 32 / 20 &\spacelena 164 / 316 &\spacelena 31 / 16 \\
\!\!\!\textsc{l 1} & \spacelena 97.5 / 89.9 &\spacelena 82 / 473 &\spacelena 62 / 11 &\spacelena 87 / 574 &\spacelena 58 / 9 \\
\!\!\!\textsc{l 2} & \spacelena 97.1 / 84.2 & \spacelena99 / 632 &\spacelena 51 / 8 &\spacelena 109 / 795 &\spacelena 47 / 6 \\

\midrule
\multicolumn{5}{l}{\!\!\!\bf MLP-1, h=100 }\\
\!\!\!\textsc{l 0} & \spacelena 93.7 / 95.3 & \spacelena159 / 269 &\spacelena 32 / 19 &\spacelena 163 / 374 &\spacelena 31 / 14 \\
\!\!\!\textsc{l 1} & \spacelena 97.6 / 86.4 &\spacelena 82 / 525 &\spacelena 62 / 10 &\spacelena 87 / 683 &\spacelena 58 / 8 \\
\!\!\!\textsc{l 2} & \spacelena 97.1 / 80.0 &\spacelena 100 / 731 &\spacelena 51 / 7 &\spacelena 109 / 905 &\spacelena 47 / 6 \\

\midrule
\multicolumn{5}{l}{\!\!\!\bf MLP-1, h=50}\\
\!\!\!\textsc{l 0} &\spacelena 93.7 / 92.7 &\spacelena 159 / 336 &\spacelena 32 / 15 &\spacelena 164 / 438 &\spacelena 31 / 11 \\
\!\!\!\textsc{l 1} &\spacelena 97.6 / 82.0 &\spacelena 82 / 648 &\spacelena 62 / 8 &\spacelena 90 / 790 &\spacelena 56 / 7 \\
\!\!\!\textsc{l 2} &\spacelena 97.2 / 75.0 &\spacelena 100 / 875 &\spacelena 51 / 6 &\spacelena 114 / 1016 &\spacelena 45 / 5 \\

\bottomrule
\end{tabular}
\caption{Experimental results; shown in pairs: linguistic task / control task. Codelength is measured in kbits (variational codelength is given in equation (\ref{eq:variational_code}), online -- in equation (\ref{eq:online_code})). $h$ is the probe hidden layer size.}
\label{tab:control_results_settings}
\end{table}

\subsection{Random seeds: control task}

\begin{figure}[h!]
    \centering
     {\includegraphics[scale=0.4]{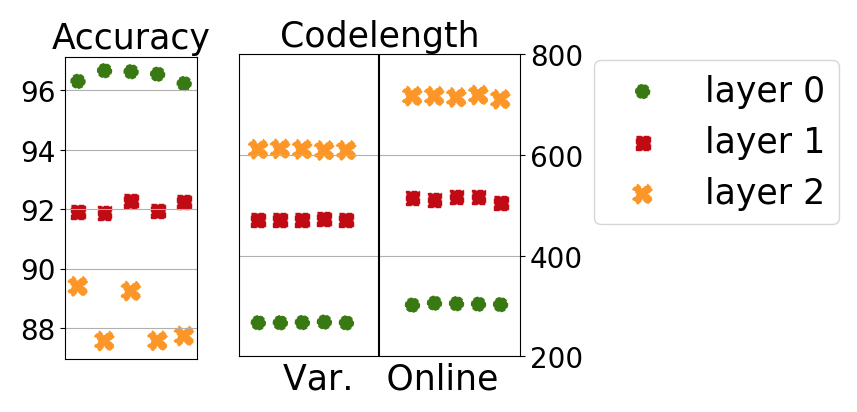}}
    \caption{Results for 5 random seeds, control task (default setting: MLP-2, $h=1000$). }
    \label{fig:control_seeds_control}
\end{figure}

Results are shown in Figure~\ref{fig:control_seeds_control}.

\section{Description Length and Random Models}

\begin{table}[h!]
\centering
\begin{tabular}{lccc}
\toprule
& & \bf Accuracy & \bf Final probe \\
\midrule

\multicolumn{2}{l}{\!\!\bf layer 0}\\
& \!\!\!\!\!\!base & 91.31 & 728-31-154 \\
\midrule

\multicolumn{2}{l}{\!\!\bf layer 1}\\
& \!\!\!\!\!\!base & 97.7 & 878-42-172 \\
& \!\!\!\!\!\!random & 96.76 & 876-50-228 \\
\midrule

\multicolumn{2}{l}{\!\!\bf layer 2}\\
& \!\!\!\!\!\!base & 97.32 & 872-50-211 \\
& \!\!\!\!\!\!random & 96.76 & 929-47-229 \\
\bottomrule
\end{tabular}
\caption{Pruned architecture of a trained variational probe, Part of Speech (starting probe: 1024-256-256). }
\label{tab:jiant_pos_bayes_pruned_architecture}
\end{table}

\begin{table}[h!]
\centering
\begin{tabular}{lccc}
\toprule
& & \bf Accuracy & \bf Final probe \\
\midrule

\multicolumn{2}{l}{\!\!\bf layer 0}\\
& \!\!\!\!\!\!base & 75.61 & 976-47-242 \\
\midrule

\multicolumn{2}{l}{\!\!\bf layer 1}\\
& \!\!\!\!\!\!base & 86.01 & 1011-53-227 \\
& \!\!\!\!\!\!random & 81.35 & 1001-57-235 \\
\midrule

\multicolumn{2}{l}{\!\!\bf layer 2}\\
& \!\!\!\!\!\!base & 84.36 & 985-61-238 \\
& \!\!\!\!\!\!random & 81.42 & 971-57-234 \\
\bottomrule
\end{tabular}
\caption{Pruned architecture of a trained variational probe, constituent labeling (starting probe: 1024-256-256). }
\label{tab:jiant_nonterm_bayes_pruned_architecture}
\end{table}

\begin{table}[h!]
\centering
\begin{tabular}{lccc}
\toprule
& & \bf Accuracy & \bf Final probe \\
\midrule

\multicolumn{2}{l}{\!\!\bf layer 0}\\
& \!\!\!\!\!\!base & 80.11 & (423+356)-36-119 \\
\midrule

\multicolumn{2}{l}{\!\!\bf layer 1}\\
& \!\!\!\!\!\!base & 92.3 & (682+565)-38-85 \\
& \!\!\!\!\!\!random & 89.86 & (635+548)-40-98 \\
\midrule

\multicolumn{2}{l}{\!\!\bf layer 2}\\
& \!\!\!\!\!\!base & 90.6 & (581+422)-42-104 \\
& \!\!\!\!\!\!random & 89.96 & (646+538)-38-94 \\
\bottomrule
\end{tabular}
\caption{Pruned architecture of a trained variational probe, dependency labeling (starting probe: (1024+1024)-512-256). }
\label{tab:jiant_dep_bayes_pruned_architecture}
\end{table}

\begin{table}[h!]
\centering
\begin{tabular}{lccc}
\toprule
& & \bf Accuracy & \bf Final probe \\
\midrule

\multicolumn{2}{l}{\!\!\bf layer 0}\\
& \!\!\!\!\!\!base & 91.7 & 450-16-36 \\
\midrule

\multicolumn{2}{l}{\!\!\bf layer 1}\\
& \!\!\!\!\!\!base & 94.95 & 509-16-35 \\
& \!\!\!\!\!\!random & 93.36 & 551-18-36 \\
\midrule

\multicolumn{2}{l}{\!\!\bf layer 2}\\
& \!\!\!\!\!\!base & 94.93 & 527-17-41 \\
& \!\!\!\!\!\!random & 93.57 & 536-18-34 \\
\bottomrule
\end{tabular}
\caption{Pruned architecture of a trained variational probe, named entity recognition (starting probe: 1024-256-256). }
\label{tab:jiant_ner_bayes_pruned_architecture}
\end{table}

\begin{table}[h!]
\centering
\begin{tabular}{lccc}
\toprule
& & \bf Accuracy & \bf Final probe \\
\midrule

\multicolumn{2}{l}{\!\!\!\bf layer 0}\\
& \!\!\!\!\!\!base & 79.1 & (567+754)-46-158 \\
\midrule

\multicolumn{2}{l}{\!\!\!\bf layer 1}\\
& \!\!\!\!\!\!base & 90.25 & (709+937)-48-140 \\
& \!\!\!\!\!\!random & 86.59 & (678+857)-55-148 \\
\midrule

\multicolumn{2}{l}{\!\!\!\bf layer 2}\\
& \!\!\!\!\!\!base & 88.5 & (601+863)-52-142 \\
& \!\!\!\!\!\!random & 86.34 & (744+889)-53-151 \\
\bottomrule
\end{tabular}
\caption{Pruned architecture of a trained variational probe, semantic role labeling (starting probe: (1024+1024)-512-256). }
\label{tab:jiant_srl_bayes_pruned_architecture}
\end{table}

\begin{table}[h!]
\centering
\begin{tabular}{lccc}
\toprule
& & \bf Accuracy & \bf Final probe \\
\midrule

\multicolumn{2}{l}{\!\!\bf layer 0}\\
& \!\!\!\!\!\!base & 88.87 & (358+352)-16-20 \\
\midrule

\multicolumn{2}{l}{\!\!\bf layer 1}\\
& \!\!\!\!\!\!base & 91.6 & (497+492)-20-22 \\
& \!\!\!\!\!\!random & 90.35 & (363+357)-23-21 \\
\midrule

\multicolumn{2}{l}{\!\!\bf layer 2}\\
& \!\!\!\!\!\!base & 90.29 & (519+505)-18-19 \\
& \!\!\!\!\!\!random & 90.45 & (375+377)-21-21 \\
\bottomrule
\end{tabular}
\caption{Pruned architecture of a trained variational probe, coreference resolution (starting probe: (1024+1024)-512-256). }
\label{tab:jiant_coref_bayes_pruned_architecture}
\end{table}

\begin{table}[h!]
\centering
\begin{tabular}{lccc}
\toprule
& & \bf Accuracy & \bf Final probe \\
\midrule

\multicolumn{2}{l}{\!\!\!\bf layer 0}\\
& \!\!\!\!\!\!base & 48.77 & (138+137)-10-14 \\
\midrule

\multicolumn{2}{l}{\!\!\!\bf layer 1}\\
& \!\!\!\!\!\!base & 71.07 & (116+178)-16-17 \\
& \!\!\!\!\!\!random & 60.73 & (168+135)-15-15 \\
\midrule

\multicolumn{2}{l}{\!\!\!\bf layer 2}\\
& \!\!\!\!\!\!base & 71.59 & (123+164)-14-18 \\
& \!\!\!\!\!\!random & 60.69 & (167+125)-13-15 \\
\bottomrule
\end{tabular}
\caption{Pruned architecture of a trained variational probe, relation classification (starting probe: (1024+1024)-512-256). }
\label{tab:jiant_relsem_bayes_pruned_architecture}
\end{table}

\end{document}